\documentclass{ifacconf}

\usepackage{graphicx}      
\usepackage{amsmath,amssymb,amsfonts} 

\makeatletter
\let\old@ssect\@ssect 
\makeatother

\usepackage{natbib}
\usepackage{hyperref}

\makeatletter
\def\@ssect#1#2#3#4#5#6{%
  \NR@gettitle{#6}
  \old@ssect{#1}{#2}{#3}{#4}{#5}{#6}
}
\makeatother

\begin{document}
\begin{frontmatter}

\title{2.5D Mapping, Pathfinding and Path Following For Navigation Of A Differential Drive Robot In Uneven Terrain\thanksref{footnoteinfo}}

\thanks[footnoteinfo]{This work was supported by Russian Science Foundation, project №22-21-00716.\\This is a preprint of the paper accepted to IFAC SYROCO'21/22}

\author{Stepan Dergachev $^{*, **}$} 
\author[First]{Kirill Muravyev} 
\author{Konstantin Yakovlev $^{*, **}$}

\address[First]{Federal Research Center ``Computer Sciences and Control'' of Russian Academy of Sciences, Moscow, Russia (e-mail:\{dergachev, muraviev, yakovlev\}@isa.ru).}
\address[Second]{National Research University Higher School of Economics, Moscow, Russia}

\begin{abstract}                
Safe navigation in uneven terrains is an important problem in robotic research. In this paper we propose a 2.5D navigation system which consists of elevation map building, path planning and local path following with obstacle avoidance. For local path following we use Model Predictive Path Integral (MPPI) control method. We propose novel cost-functions for MPPI in order to adapt it to elevation maps and motion through unevenness. We evaluate our system on multiple synthetic tests and in a simulated environment with different types of obstacles and rough surfaces.
\end{abstract}


\end{frontmatter}

\section{Introduction}

Autonomous navigation of a mobile robot is an important problem for both practitioners and researchers. A typical approach to navigation relies on the usage of 2D occupancy grid maps for path planning and path following. 2D occupancy grids are convenient for path planning and contain detailed information about the obstacles around a robot. However, when a robot operates in indoor or industrial environments, it often faces up different uneven surfaces (ramps, stairs, road borders, rough terrains, rubbish on a path, etc.). These surfaces may be traversable for robot but are typically marked as obstacles on 2D occupancy grid. So, 2D navigation in such environments becomes complicated. This problem may be solved by building detailed 3D maps, but path planning in them takes significant computational efforts that may be critical for low-power onboard computer.

Another approach to navigation in uneven environments is the usage of height maps (2.5D maps). These maps are represented as a grid where each cell encodes elevation of corresponding area of a surface. Optionally, 2.5D grid maps may contain additional information like uncertainty. 

Elevation information from 2.5D maps is widely used to estimate traversability of rough terrain. This traversability may be computed as binary value (cells can be classified as traversable or untraversable), or as a cost-function (how difficult to traverse or how dangerous the region is). These traversability values can be utilized by path planning and path following algorithms to compute both effective and safe trajectory through uneven surface.

In most navigation systems, robot is moved along the path planned by global path planner in a global map. In case of uneven terrains or dynamic objects, there may be small unmapped obstacles (pits, hillocks, stones, garbage etc.) which can impede robot motion. In order to avoid these small obstacles, local path correction is typically applied. One of the approaches to local path correction is usage of Model Predictive Control (MPC)-based methods, partially Model Predictlve Path Integral (MPPI) control \citep{williams2016aggressive}. These methods compute sequence of controls for robot as a weighted average over trajectories sampled from system dynamics. The weights for averaging are taken from cost-function values along trajectories. MPPI control based methods are widely used, but mainly with 2D environment representations.


In this work we adapt MPPI control method to uneven terrains using 2.5D elevation maps. We use local elevation map as an input for MPPI algorithm. MPPI is guided by terrain traversability values computed by this elevation map. These traversability values are computed from slope steepness, surface roughness and other parameters. We carry out wide experimental evaluation of our navigation pipeline in various simulated environments. We integrate our pipeline with ROS2\footnote{https://docs.ros.org/en/foxy/} framework widely used by robotic researchers and implement it as a part of the Nav2 software stack \citep{macenski2020marathon2}.

\section{Related Works}

\subsection{Elevation Mapping}

Building height maps for robot navigation is not new. \citep{gutmann2005floor} proposed building of height map divided into floor and obstacles for humanoid robot navigation. In~\citep{mastalli2017trajectory} height grid maps are constructed for safe and efficient motion of four-legged robots. \citep{de2017skimap} propose construction of global 2.5D elevation map as a projection of global 3D map for Micro Aerial Vehicle (MAV) navigation. 

A universal, robust and efficient local elevation mapping method was proposed by~\citep{fankhauser2018probabilistic}. In their work, local robot-centric elevation map is aggregated from range and pose sensor measurements with Kalman filter correction and refined via probabilistic fusion. A source code of this method is commonly available\footnote{https://github.com/ANYbotics/elevation\_mapping}. In our work, we take this method as a baseline and modify it for efficient path planning and following.

\subsection{Path Planning}

One of the most common approaches to robot trajectory planning is to represent the environment as a graph and find paths in this graph. The vertices of the graph are considered as valid positions or states of the robot, and an edge between vertices means a transition between corresponding states. The vertices can form a regular grid~\citep{elfes1989using, yap2002grid}, or be generated randomly~\citep{kavraki1996probabilistic}. One of the most common algorithms for graph path finding is ~\citep{dijkstra1959note}, but currently heuristic search approaches such as $A^*$~\citep{hart1968formal}, $Theta^*$~\citep{daniel2010theta}, $D^*$-Lite~\citep{koenig2005fast} are widely used in this problem. 

On the other hand, there is also a group of probabilistic approaches to trajectory planning, such as RRT~\citep{lavalle2000rapidly} or RRT*~\citep{karaman2011sampling}. In this case, the robot states are generated by the algorithm during the planning process by sampling the workspace. For each state its validity is checked on the fly, as well as the feasibility of the transition from one of the already sampled states.

\subsection{Path Following in Uneven Terrain}

Safe and efficient robot motion in rough terrains is a challenging problem, and many works try to address it. \citep{shimoda2007high} and~\citep{raja2015new} navigate in rough terrains using 2.5D grid maps and potential field methods. \citep{tahirovic2011roughness} used well-known RRT algorithm with roughness-based cost-functions which can also be applied for Model Predictive Control (MPC) methods. 

MPC methods were used widely for rough terraing locomotion, see~\citep{tahirovic2010passivity, fan2021step, buyval2019architecture}. \citep{fan2021step} solved risk-aware MPC problem to minimize traversability cost in uneven terrains and simultaneously minimizing risk of collision or tip-over. \citep{buyval2019architecture} used MPPI control method for off-road navigation of autonomous truck.

A separate subtask is the evaluation of the trajectories at uneven terrain. The simplest methods use absolute height differences~\citep{shang2021complete} or the Euclidean distance in 3D space between waypoints of the trajectory~\citep{gu2011path}. Other methods, e.g.~\citep{sock2014probabilistic}, allow to estimate the traversability of the individual areas of the environment, on the basis of which evaluation of trajectories can be carried out.More complex methods estimate the slope of the robot in the position, or the degree of surface unevenness nearby of the position:~\citep{tahirovic2011roughness}, \citep{gu2008rapid}, \citep{ye2004method}.

In our work, we use Model Predictive Path Integral (MPPI) control method for local path following and examine different cost-functions. These cost-functions consider traversability, roughness, and slope angle along the path.

\section{Problem Statement}

Consider a differential drive robot tasked to move to the target location through the uneven terrain. A robot is equipped with an odometry sensor which provides 6-DoF robot's pose and its speed, and an RGB-D sensor which provides point clouds from robot's front view.

At each moment $t$, the robot has state $s_t = (\mathbf{p_t, q_t, v_t, w_t}),$ where $\mathbf{p_t} \in \mathbb{R}^3$ is the position, $\mathbf{q_t} \in \mathbb{R}^3$ is the orientation, $\mathbf{v_t} \in \mathbb{R}^3$ is the transitional velocity, and $\mathbf{w_t} \in \mathbb{R}^3$ is the angular velocity. All these are vectors.

The point cloud $P_t = \{ (x_i, y_i, z_i);\ i = 1, \dots, N \},$ is a set of $N$ 3D points observed by robot's RGB-D sensor, where $i$-th point has coordinates $(x_i, y_i, z_i)$ in the sensor-related coordinate system.

The robot state is changing with the law $s_{t+1} = f(s_t, \mathbf{u_t})$, where $\mathbf{u_t}$ is a control (input) vector at time step $t$ (e.g. a vector of linear and angular acceleration which are sent to the robot's controller).

The navigation system is also provided with the goal state $s_G = (\mathbf{p_g}, \mathbf{q_g}, \mathbf{v_g}=0, \mathbf{w_g}=0)$. Its output at step $t$ is a control input $\mathbf{u_t}$ which is sent to the robot's controller in order to reach the goal:

\[\pi(s_t; P_1, \dots, P_t) = \mathbf{u_t}\]

Our aim is to build such function $\pi$ which moves a robot to goal pose with spatial and angular deviation under some thresholds:

$$\pi(s_T, P_1, \dots, P_T) = \mathbf{u_T};\ \  f(s_T, \mathbf{u_T}) \approx s_G,$$

where $(\mathbf{p_a}, \mathbf{q_a}, \mathbf{v_a}, \mathbf{w_a}) \approx (\mathbf{p_b}, \mathbf{q_b}, \mathbf{v_b}, \mathbf{w_b})$ means that $||\mathbf{p_a} - \mathbf{p_b}|| < \alpha$ and $||\mathbf{q_a} - \mathbf{q_b}|| < \beta$ where $\alpha$ and $\beta$ are the pre-defined thresholds.

\section{Navigation System}

\subsection{Overview}
We propose a fully autonomous navigation system for uneven environments. The system consists of the following modules:

\begin{itemize}
    \item \textbf{Elevation mapping} module takes RGBD and odometry sensor data and builds a height map of the robot's surroundings.
    \item \textbf{Path planning} module builds a global plan to reach the goal. Its inputs are: the robot's odometry data, target state, and the height map.
    \item \textbf{Path following} module follows the trajectory suggested by the path planner and performs local maneuvers to avoid obstacles and severe roughnesses. It takes a robot's odometry data, global path, and 2.5D map built by the mapping module as an input.
\end{itemize}

All the modules run in real time as independent threads. Elevation mapping and path planning modules run at the low frequency (about 1 Hz). The path follower runs with a higher frequency (about 10 Hz) to maintain robot's motion and quickly react to surface roughness changes.

\subsection{Mapping}
For elevation mapping, we use the approach of \citep{fankhauser2018probabilistic}. It is an open-source 2.5D mapping solution which updates height map from robot's motion and point clouds using Kalman filter and probabilistic vicinity-based map fusion. During map fusion, height and variance of each cell are updated from cumulative distribution function (CDF) calculated by points of confidence ellipse around the cell. Such probabilistic fusion lets us increase map accuracy and consistency and significantly reduce its uncertainty. Our tests in simulation with noised odometry show that this method is robust to odometry sensor noise and is able to build dense and accurate height map.

The method of~\citep{fankhauser2018probabilistic} has a serious problem for off-the-shelf navigation with standard RGB-D sensors. The problem is that RGB-D sensors usually do not observe space under the robot and directly ahead it. So, at initial stage, the robot stands in an unmapped place, and path follower is unable to move it along 2.5D map. To address this problem, we fill uncovered cells under robot's footprint guided by its spatial position and slope. At initial stage, we take larger footprint radius (0.4 meters in our experiments), and when initialization is complete, we fill map cells in smaller radius (0.25 meters), to cover occasional holes in the map.

\subsection{Path Planning and Following}

\subsubsection{Path Planning}

To create a global path, we suggest to use any of the planners available in Nav2 ROS2 package. For example, Theta* algorithm~\citep{daniel2010theta}, which is an any-angle modification of the seminal A* algorithm, can be used. This algorithm operates on a square grids, each cell of which should be marked as traversable or as untraversable.

Information about the traversability of cells can be obtained on-the-fly based on an elevation map and added on the environment map used in Navigation2 software stack. For these purposes, we propose to use the method from~\citep{sock2014probabilistic}. This method builds a probabilistic estimation $p(i, j)$ of the traversability of a cell $(i, j)$ based on the height difference between cells located at a distance of $\Delta$ along the vertical and horizontal axes.

\[ S_{h}(i, j) = \frac{|z(i, j + \Delta) - z(i, j - \Delta)|}{2\Delta\varepsilon}\]
\[ S_{v}(i, j) = \frac{|z(i + \Delta, j) - z(i - \Delta, j)|}{2\Delta\varepsilon}\]
\[ S(i, j) = \frac{S_h + S_v}{2};\ p(i, j) = e^{-\lambda \cdot S(i, j)}\]

where $z(i, j)$ is height value of cell $(i, j)$, $\varepsilon$ is the size of the cell. The adjustment of this method for a specific robot or environment is made by varying the parameters $\Delta$ and $\lambda$. A cell $(i, j)$ is considered traversable if the value $p(i, j)$ exceeds 0.5.

\subsubsection{Path Following}

The MPPI algorithm~\citep{williams2016aggressive} was chosen to follow the global path. This algorithm iteratively selects the control to follow the global path based on the estimation of a set of generated trajectories. 

At each step, the algorithm creates a set of control sequences. Each control sequence is randomly generated taking into account the kinematic and dynamic constraints of the robot. Based on each sequence of controls and the kinematic model of the robot, the robot's trajectory is built. A cost estimation $S(\pi)$ is calculated for each trajectory $\pi$. The obtained estimates are used to select the next control by applying the softmax function to the control sequences using estimated costs.

The cost estimation for individual trajectory $\pi$ is based on the following scheme:

\[S(\pi) = \sum_{i=0}^{k} \alpha_i \cdot S_i(\pi)^{\beta_i} \]

Parameters $\alpha_i$ and $\beta_i$ let us adjust the impact of individual assessment components $S_i$ on the final value. 

The proposed cost-function can be divided into two components: the general part responsible for the quality of the trajectory regardless of elevation and roughness, and the part responsible for evaluating the trajectory from the point of view of navigation in an uneven environment.

The general part includes next cost-functions:

\begin{itemize}
    \item \textbf{Distance to the trajectory} suggested by the path planner;
    \item \textbf{Distance to the goal} suggested by the path planner;
    \item \textbf{Backward motion distance} - we penalize backward motion because the robot has only forward-looking camera;
\end{itemize}

To evaluate trajectories from the point of view of navigation in an uneven environment, we suggest two different cost-functions.

The first function is based on the method described in section Path Planning. The value of the function is inversely proportional to the distance to the untraversable area closest to the trajectory. Let's denote this function as \textbf{Slope-Traversability}.

The second cost-function is denoted as \textbf{Slope-Roughness}. This function is divided into two components which are sums of inclination and roughness estimates of robot's footprint at each trajectory point. The inclination is estimated by fitting elevation data at the footprint into a plane $ax + by + c = z$ using least squares method. The roughness estimate is the standard deviation between points at the footprint and the estimated plane.

$$a_i, b_i, c_i = LSM(\{ (x_j, y_j, z_j) \in \mathcal{B}_r(\pi_i) \})$$
$$S_{slope}(\pi) = \sum\limits_{i=1}^T \arccos(\frac{1}{\sqrt{a_i^2 + b_i^2 + 1}}(a_i, b_i, -1) \cdot (0, 0, 1))$$

$$\sigma_i = \sqrt{\frac{1}{|\mathcal{B}_r(\pi_i)|} \sum\limits_{(x_j, y_j, z_j) \in \mathcal{B}_r(\pi_i)} \frac{(a_i x_j + b_i y_j + c_i - z_j)^2}{a_i^2 + b_i^2 + 1}}$$
$$S_{rough}(\pi) = \sum\limits_{i=1}^T \sigma_i$$

\section{Experimental Evaluation}

The experimental evaluation of the proposed navigation system consisted of two parts. In the first part, we performed an evaluation of the MPPI algorithm. In the second part, we carried out the tests of the entire system on a model on differential-drive robot in the Gazebo simulator~\citep{koenig2004design}.

\subsection{MPPI Experiments}

\begin{figure}[tb]
    \centering
        \includegraphics[width=0.8\columnwidth]{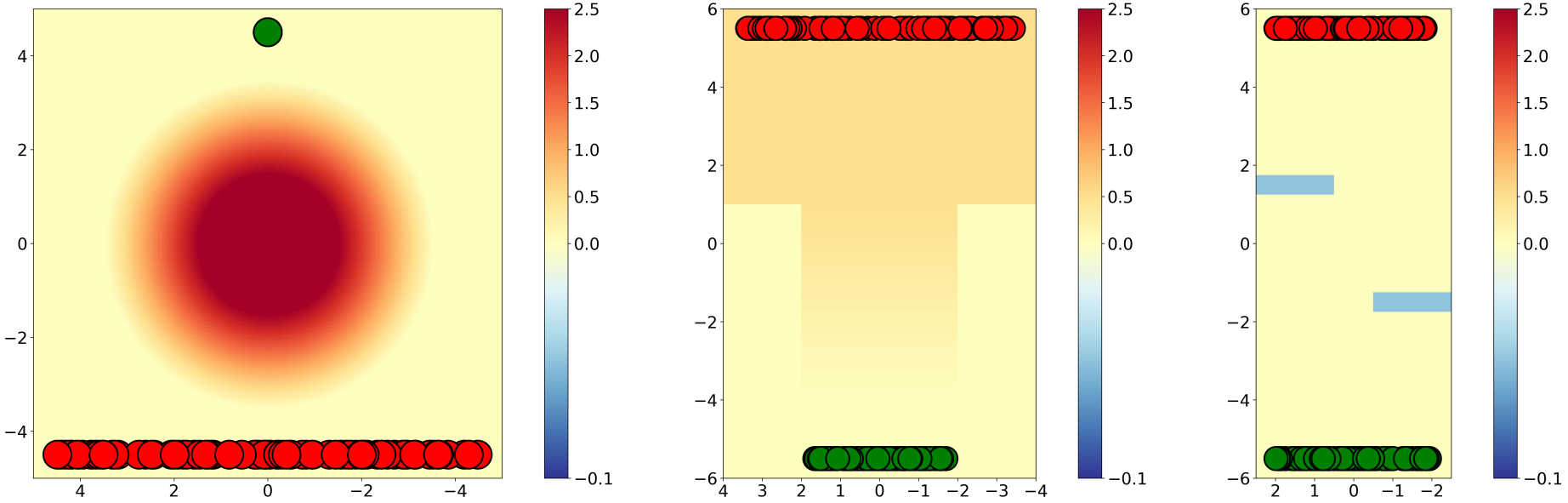}
    \caption{Maps used for experimental evaluation of our MPPI controller, with start positions (green) and goal positions (red).
    }
    \label{fig:maps}
\end{figure}

We used three different elevation maps for the experiment. The first map is a square plane of size 10x10~m with a truncated cone located in the middle, which has base diameter of 7~m, top diameter of 3~m, and height of 2.5~m. The second map consists of two surfaces having different heights and connected by an even ramp. One surface has size 7x8~m, another has size 5x8~m and height 0.5~m above first one. The ramp width is 4~m and its slope angle is about 0.1 radians. Third map is a rectangular plane of size 12~x5~m with two small rectangular pits. The size of the pits is 0.5~mx2.5~m, the depth of the pits is 0.05~m. The maps are illustrated in Fig.~\ref{fig:maps}.

We generated 100 start and goal positions for each map. These positions were distributed randomly as it is shown in Fig.~\ref{fig:maps}. In all of the test cases, when moving from start to goal position, the robot had to overcome one of three obstacles: drive around a truncated cone, drive on a ramp, drive around the pits.

The following quality metrics were used to evaluate the obtained solutions for $N$ tasks.  
\begin{itemize}
    \item $Success \, Rate$ -- the percentage of tasks for which the robot has reached the goal position (with an acceptable specified $\Delta$ deviation) in no more than a specified number of simulation steps $n_{steps}$. 
    \item $Path \, Length$ -- the average length $L$ of the robot's path $\pi$ for successfully completed tasks.
    \item $Sim \, Time$ -- the average simulation time $T$ required to reach the goal state.
\end{itemize}

Table \ref{tb:results} presents the results of the experimental launch of MPPI algorithm on considered maps with two proposed cost-functions: \texttt{Slope}-\texttt{Roughness} (SR) and \texttt{Slope}-\texttt{Traversability} (ST). The results presented in the table are calculated for tasks successfully completed by both versions of algorithm. The radius of the robot's footprint was set to 0.3~m, the maximum number of simulation steps $n_{steps}$ was 1000, and the permissible deviation $\Delta$ of the robot from the goal was 0.3~m.

\begin{table}[tb]
    \renewcommand{\arraystretch}{1.0}
    \begin{center}
        \caption{The results of experimental evaluation of MPPI algorithm with two different cost-funtions (\texttt{Slope}-\texttt{Roughness} (SR) and \texttt{Slope}-\texttt{Traversability} (ST)) at three different maps.}
        \label{tb:results}
        
        \begin{tabular}{l | c c c c c c }
         & \multicolumn{2}{c}{\texttt{Map 1}} & \multicolumn{2}{c}{\texttt{Map 2}} & \multicolumn{2}{c}{\texttt{Map 3}} \\
         & ST & SR & ST & SR & ST & SR \\ \hline
        Success Rate & 100\% & 100\% & 100\% & 100\% & 98\% & 98\%\\
        Path Length  & 11.36 & 11.18 & 11.02 & 10.99 & 10.92 & 11.26\\ 
        Sim Time & 30.51 & 29.98 & 28.52 & 28.05 & 28.38 & 31.49 \\ 
        \end{tabular}
    \end{center}
\end{table}

As can be seen in Table \ref{tb:results}, both versions of the algorithm successfully finish the vast majority of tasks. It is worth noting that both options build paths avoiding serious roughness or steep inclines. In the case of the first two maps, the algorithm with cost-function \texttt{Slope}-\texttt{Roughness} builds shorter paths on average and reaches the goal in fewer simulation steps than the algorithm with cost-function \texttt{Slope}-\texttt{Traversability}. However, for the third map, the opposite results were obtained, the paths of MPPI with \texttt{Slope}-\texttt{Roughness} cost-function are longer and require more simulation steps. This can be explained by the fact that the \texttt{Slope}-\texttt{Roughness} builds paths that avoid even small roughness, but allow passage closer to cliffs or steep inclines. 

\begin{figure}[tb]
    \centering
        \includegraphics[width=0.8\columnwidth]{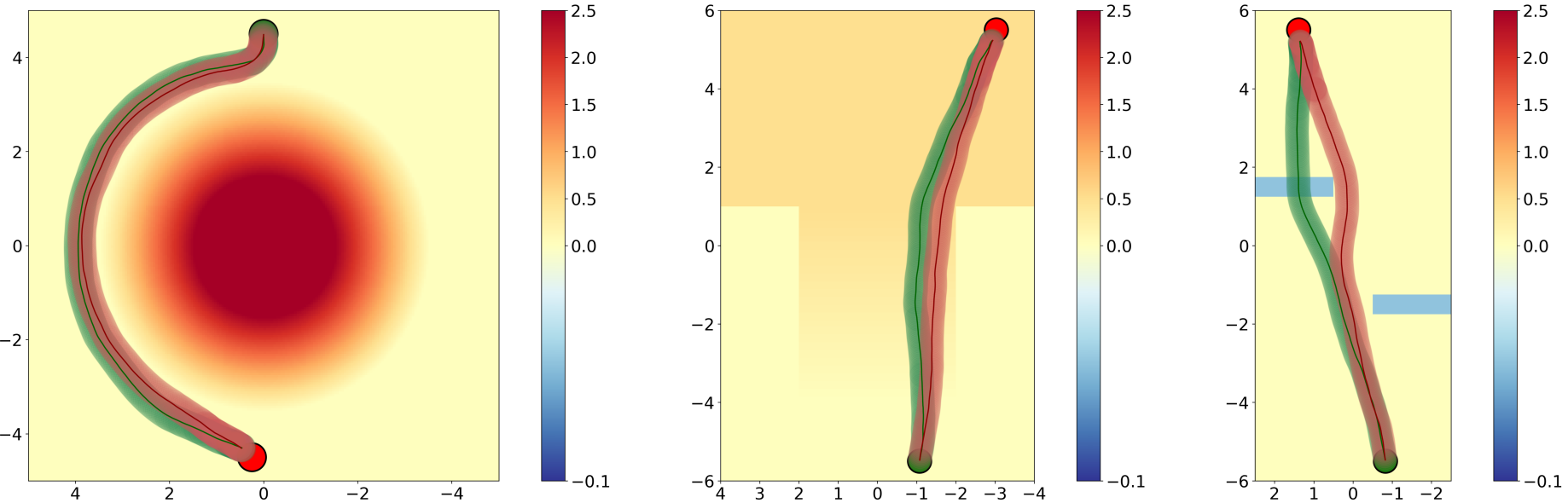}
    \caption{Example of trajectories obtained on experimental maps by two variants of the MPPI algorithm (green trajectories -- \texttt{Slope}-\texttt{Traversability}, red trajectories -- \texttt{Slope}-\texttt{Roughness}).}
    \label{fig:maps_traj}
\end{figure}

Examples of  trajectories demonstrating this behavior are shown in Fig.~\ref{fig:maps_traj}. It can be seen that the trajectories of MPPI with \texttt{Slope}-\texttt{Traversability} (marked in green) are located further from the edges of the cone on the first map and from the edge of the cliff on the second map than the trajectories of algorithm MPPI with \texttt{Slope}-\texttt{Roughness} (marked in red). But at the same time, MPPI with \texttt{Slope}-\texttt{Roughness} avoids passage through the pit, while MPPI with \texttt{Slope}-\texttt{Traversability} passes through it. 

Such behavior can be obtained due to the possibility of more fine-tuning of the cost-function, since two components are used, one of which is responsible for assessing the roughness of the surface, and the other for the slope.

\subsection{Experiments in Simulation}

\begin{figure}[t]
    \centering
    \includegraphics[width=0.7\columnwidth]{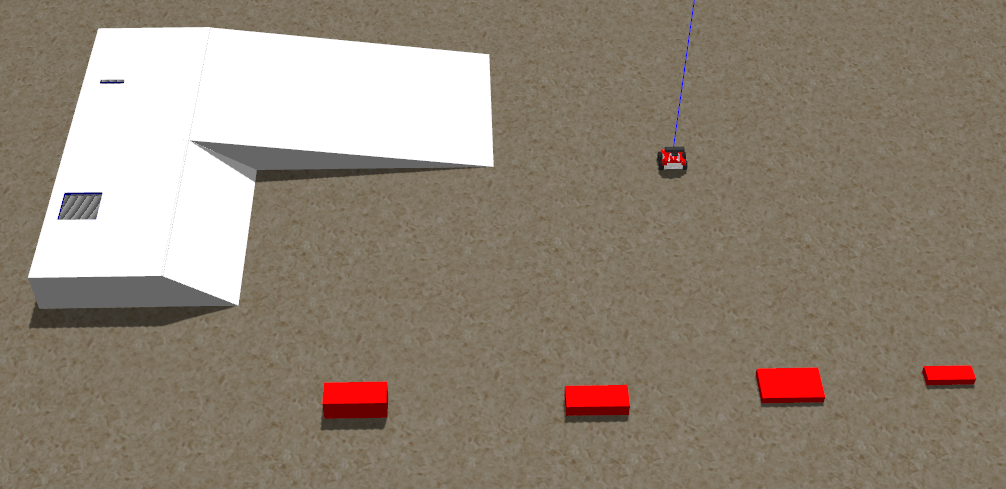}
    \caption{The environment for simulation experiment.}
    \label{fig:gazebo_environment}
\end{figure}

For simulation experiments, we created a world with different types of unevenness in Gazebo simulator. The world contains a platform of size 3x1~m and height 0.4~m, with two ramps leading to this platform. One ramp is of low-grade slope (0.12 rad), and the other ramp is of steep slope (0.59~rad). There is a small quadratic pit and small ledge on the platform. The pit has size of 30x30~cm and depth of 2.5~cm. The ledge has size of 3x20~cm and height of 1.5~cm. Also this world contains small bricks of different sizes on the ground. The height of the bricks vary from 8~cm to 27~cm, and the width is about 40~cm. Proposed world is shown in Fig. \ref{fig:gazebo_environment}.

We used a model of four-wheeled differential drive robot. This robot had wheel base of 11 cm, length of 20 cm, width of 22 cm, and clearance of 2 cm. It could easily traverse through the low-slope ramp, and tipped over on the high-slope ramp. All the bricks were impassable for this robot, but it was able to traverse through the ledge and the pit on the platform (with significant swing while entering and exiting the pit). Maximal linear speed of the robot was 0.5 m/s, and maximal rotation speed was 1.3 rad/s.

For our tests, we created a route that was divided into ten straight-line segments. First, it entered the platform using the low-slope ramp. Next, it went to the opposite end of the platform through the ledge and the pit. Then, it went down to the ground crossing a small part of the steep ramp. And finally, it had several loops on the ground, crossing the bricks and returning to the start point. A straight line connecting the robot's current position and the next waypoint on a route was used as a reference path for the path follower. The task of the path follower was to go along the route and perform appropriate detours when route segments pass through an obstacle or an undesirable roughness. 

The point clouds for the elevation mapping module came from RGBD camera model with resolution 160x120. The experiments were running on a PC with Intel i9-11900KF CPU (8 cores, 5.1 GHz). Path follower was running on a single thread, and other threads were occupied by simulation and elevation mapping. Frequency of controller was set to 10 Hz. 

\begin{figure}[t]
    \centering
    \includegraphics[width=0.75\columnwidth]{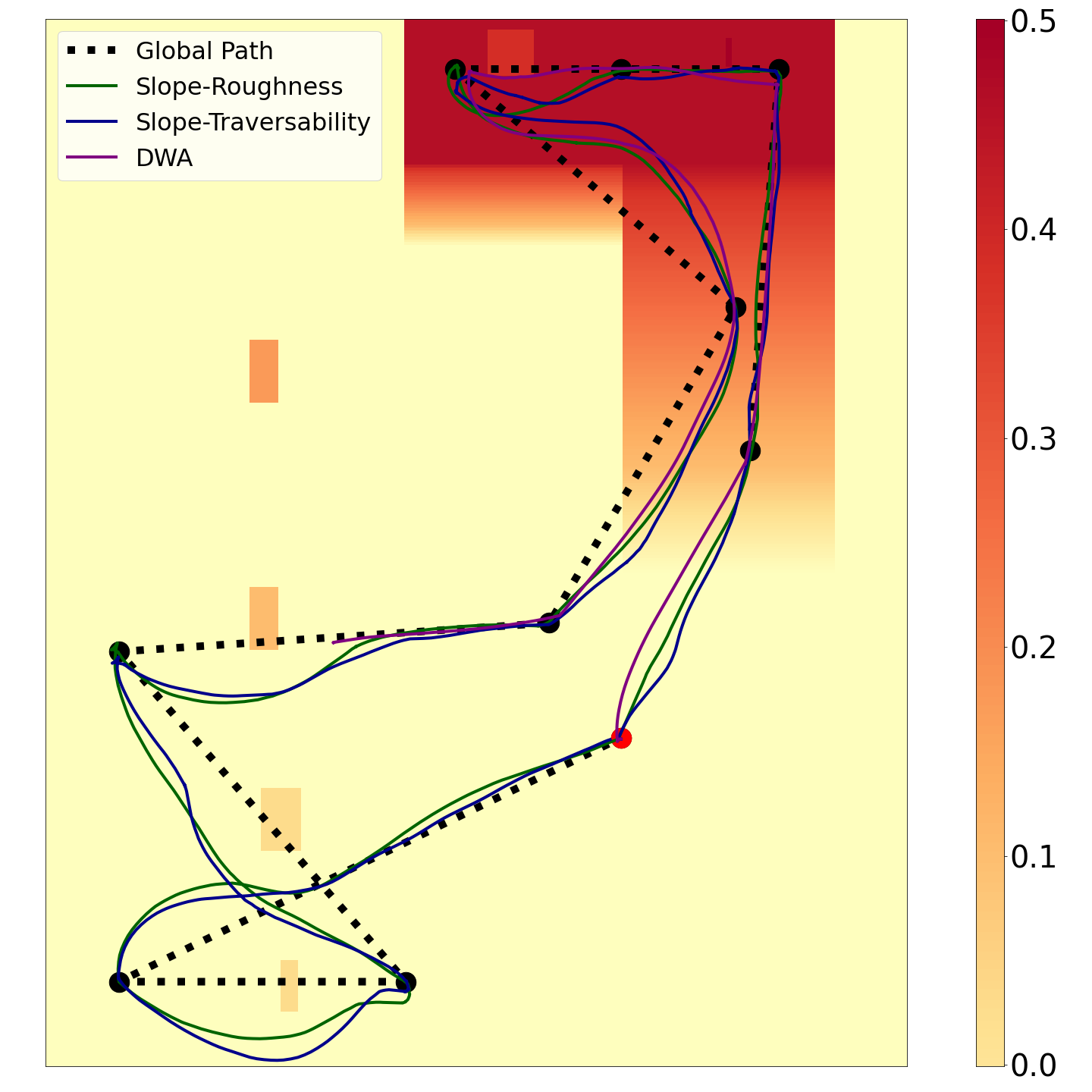}
    \caption{The route for simulation experiments (black dashed lines) and trajectories traveled by the robot with DWA (purple line),  MPPI with \texttt{Slope-Roughness} (green line) and MPPI with \texttt{Slope-Traversability} (dark blue line). The map color encodes surface elevation.}
    \label{fig:gazebo_results}
\end{figure}

The tests were carried out using MPPI algorithm with \texttt{Slope-Roughness} and \texttt{Slope-Traversability} cost-functions and also ROS2 DWB~Controller, which is based on well-known Dynamic Window Approach \citep{fox1997dynamic}. The maximum permissible deviation from the goal was set as 0.15~m and 0.25~radian. For the MPPI algorithm the number of the simulation steps was 30, the number of trajectory samples was 300, the footprint radius was set to 0.2~m. As an occupancy grid for DWB~Controller, an elevation map binarized by \texttt{Slope-Traversability} was used. Moreover, unknown area of the elevation map was also considered as untraversable. In order for the robot not to get close to the obstacles, they were inflated with an inflating radius of 0.25~m. The resultant trajectories of MPPI and Dynamic Window Approach (DWA) are visualized in Fig.~\ref{fig:gazebo_results}. The video is available at \url{https://youtu.be/LGhKaxnL8xA}.

\begin{table}[tb]
    \renewcommand{\arraystretch}{1.0}
    \begin{center}
        \caption{Results of experimental evaluation in Gazebo with \texttt{Slope-Roughness} and \texttt{Slope-Traversability} cost-functions.}
        \label{tb:results_gazebo}
        
        \begin{tabular}{l | c | c}
         & Slope-Roughness & Slope-Traversability\\
         \hline
         Path Length, m & 26.2 & 25.9\\
         Travel Time, s & 114 & 119\\
         Step Time, ms & 21 & 90\\
        \end{tabular}
    \end{center}
\end{table}

As a result, with both cost-functions, the MPPI successfully completed the task and avoided collision with the bricks and falling down the steep ramp. The DWA algorithm overcame the ramp and the platform stages, but could not build a trajectory to bypass the brick, since this would require a significant deviation from the reference path. Also, with both cost-functions, the robot moved around the pit instead of traversing it, but DWA dit not. At the same time, all algorithms moved robot through the ledge, because elevation mapping module build flat surface in place of the ledge.

For both cost-functions, traveled time and distance were measured, as well as average time to compute controls for one step. The results are shown in Table~\ref{tb:results_gazebo}. \texttt{Slope-Traversability} completed the mission a little slower but with a slightly less travel distance than \texttt{Slope-Roughness}, and took much more time for control computation (90 ms per step compared to 21 ms).

\section{Conclusion and Future Work}

In this work we proposed a pipeline for navigation of a differential drive robot in uneven terrain, which consists of elevation mapping, path planning, and path following modules. We tested the proposed pipeline on different synthetic maps and evaluated it on a simulated robot. Both synthetic and simulation experiments showed success rate close to 100\% and proved that proposed path following method is able to safely navigate through uneven terrain avoiding obstacles and large roughnesses.

In the future, we plan to create a more efficient implementation of the MPPI algorithm by parallelizing computations using CUDA/OpenCL toolkits. Another area of future work is increasing the robustness of MPPI and adapting this approach to a larger class of dynamic systems by using the ideas described in~\citep{gandhi2021robust} and~\citep{williams2017information}. Finally, conducting experiments on a real robot is also a perspective direction of future work.

\bibliography{ifacconf}          
                                                   
\end{document}